\PassOptionsToPackage{numbers}{natbib}
\documentclass{article}

\usepackage[preprint]{neurips_2024}

\usepackage[utf8]{inputenc} 
\usepackage[T1]{fontenc}    
\usepackage{hyperref}       
\usepackage{url}            
\usepackage{booktabs}       
\usepackage{amsfonts}       
\usepackage{nicefrac}       
\usepackage{microtype}      
\usepackage{xcolor}         

\usepackage{natbib}

\usepackage{amsmath}
\usepackage{enumitem}
\usepackage{algorithm}
\usepackage{algpseudocode}
\usepackage{xcolor}
\usepackage{tabularx}
\usepackage{graphicx}
\usepackage{authblk}
\usepackage{caption}

\title{DeepDistill: Enhancing LLM Reasoning Capabilities via Large-Scale Difficulty-Graded Data Training}

\author{\normalfont
  Xiaoyu Tian, \quad Sitong Zhao, \quad Haotian Wang, \quad Shuaiting Chen,\\[0.3em]
  Yiping Peng,\quad Yunjie Ji, \quad Han Zhao,\quad Xiangang Li
}

\begin{document}

\affil{
    \raisebox{-0.4em}{\includegraphics[height=1.5em]{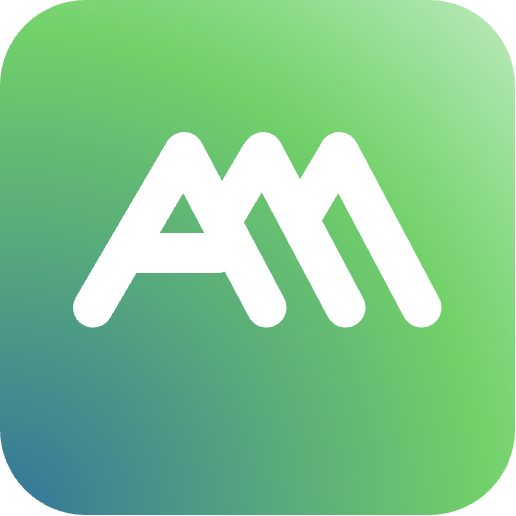}}
    \hspace{0.2em}a-m-team
}
\date{}

\maketitle

\vspace{-1.5em}
\begin{abstract}
\noindent Although large language models (LLMs) have recently achieved remarkable performance on various complex reasoning benchmarks, the academic community still lacks an in-depth understanding of base model training processes and data quality. To address this, we construct a large-scale, difficulty-graded reasoning dataset containing approximately 3.34 million unique queries of varying difficulty levels and about 40 million distilled responses generated by multiple models over several passes. Leveraging pass rate and Coefficient of Variation (CV), we precisely select the most valuable training data to enhance reasoning capability. Notably, we observe a training pattern shift, indicating that reasoning-focused training based on base models requires higher learning rates for effective training. Using this carefully selected data, we significantly improve the reasoning capabilities of the base model, achieving a pass rate of 79.2\% on the AIME2024 mathematical reasoning benchmark. This result surpasses most current distilled models and closely approaches state-of-the-art performance. We provide detailed descriptions of our data processing, difficulty assessment, and training methodology, and have publicly released all datasets and methods to promote rapid progress in open-source long-reasoning LLMs. The dataset is available at: \href{https://huggingface.co/datasets/a-m-team/AM-DeepSeek-Distilled-40M}{https://huggingface.co/datasets/a-m-team/AM-DeepSeek-Distilled-40M}
\end{abstract}

\vspace{-1em}
\begin{figure}[ht]
    \centering
    \includegraphics[width=0.94\linewidth]{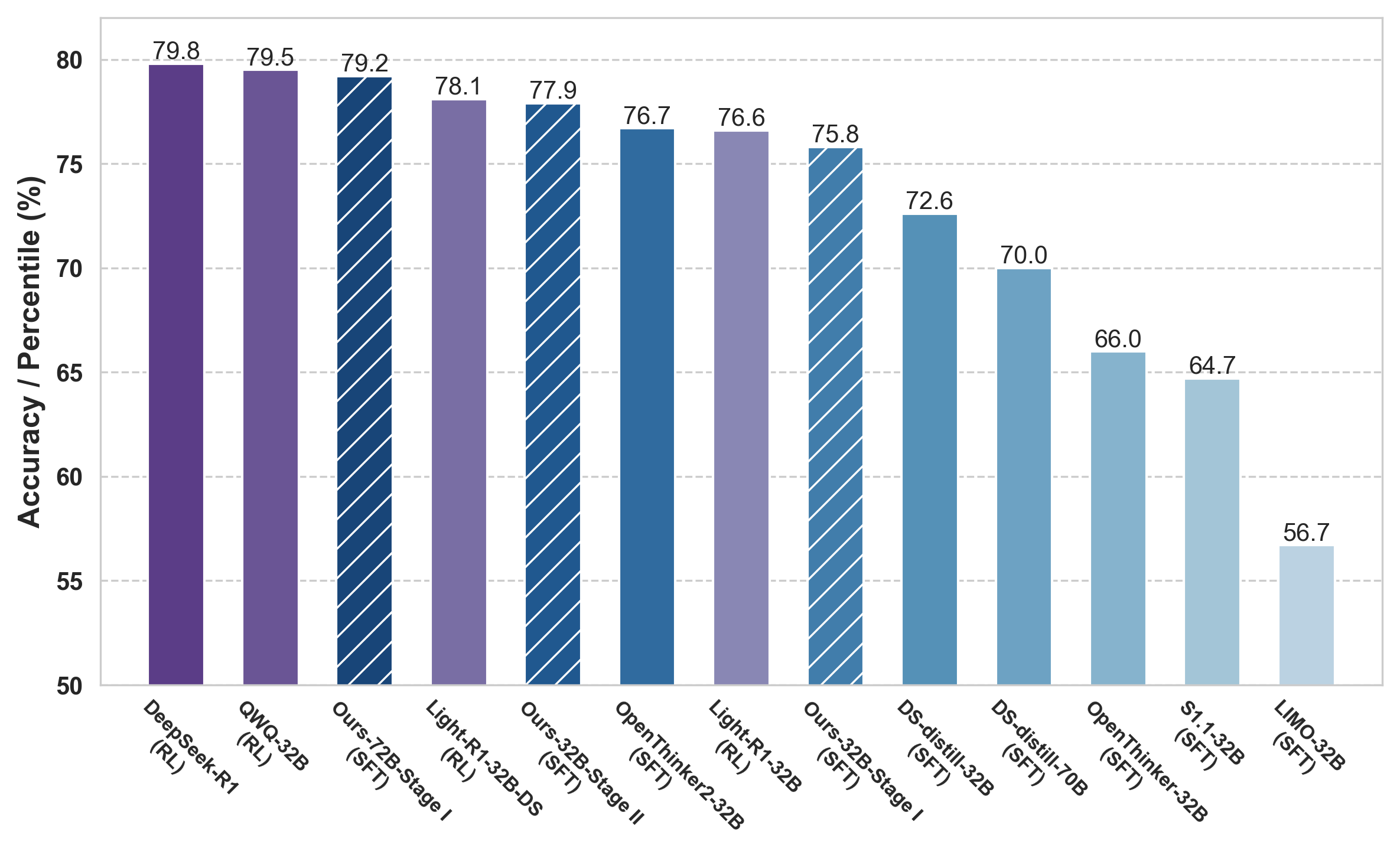}
    \vspace{-1.5em}
    \caption{Benchmark performance of open-source models on AIME2024.}
    \label{fig:hist_benchmark}
\end{figure}

\section{Introduction}

Large language Models (LLMs) have recently achieved remarkable results on complex reasoning tasks. Proprietary systems like OpenAI’s O-Series\citep{OpenAI2024} demonstrate exceptional performance on challenging benchmarks, from competitive programming to science and math.

For instance, GPT‑4o solves only ~13\% of problems on the rigorous AIME2024 Math Competition\citep{maa_aime_2024}, whereas OpenAI’s specialized o1 model—trained to reason before answer—achieves about 73\% on the same test. Similarly, reinforcement learning-driven models like DeepSeek-R1\citep{deepseekai2025deepseekr1incentivizingreasoningcapability} have attained remarkable performance, ranking among the top open-source reasoning models.

These successes underscore that scaling by extending the length of Chain-of-thought reasoning process\citep{wei2023chainofthoughtpromptingelicitsreasoning, snell2024scalingllmtesttimecompute, wu2025inferencescalinglawsempirical, tian2025think} and training with appropriate strategies can substantially improve model capabilities: OpenAI’s O-Series and DeepSeek-R1 was trained via large-scale reinforcement learning on reasoning tasks. Recent studies\citep{zhao20251, wen2025light, ji2025difficulty} have showed that carefully selecting data by well-defined difficulty levels can markedly improve training. 

Yet, despite these breakthroughs, the academic community still lacks a deep understanding of how to train base models effectively for long-form reasoning. It remains unclear how to systematically endow a base model with advanced reasoning ability through Supervised Fine-Tuning (SFT).

One key challenge is identifying and leveraging the right training data to teach reasoning. Complex reasoning problems vary widely in difficulty and form (mathematical, code generation, logical puzzles, etc.), and not all examples are equally useful for learning. Previous work\citep{li2023quantity, zhang2025d3} typically rely on a single model’s scoring or judgement to define problem difficulty or quality. This single-judge strategy is inherently biased and highly sensitive to the model used, the definition of difficulty, and the prompting method. A problem that one model finds difficult might be easy for another, and prompt phrasing can drastically alter a model’s perceived success. As a result, using a single model’s perspective can lead to less generalizable learning outcomes – models may overfit to the evaluator’s idiosyncrasies rather than gaining a broad problem-solving skill.

To address these limitations, we propose a large-scale difficulty-graded reasoning dataset and difficulty-aware reasoning data selection strategy to improve long-form reasoning in LLM training. Instead of trusting any single model to label which examples are hard or useful, we leverage a multi-pass distillation approach across different capabilities of models. Concretely, we construct a large-scale dataset where each query is accompanied by numerous solution attempts from models of varying strengths. 

Our contributions are twofold:

\begin{itemize}

\item A large-scale difficulty-graded reasoning dataset – We present a dataset of 3.34 million unique reasoning queries, each paired with rich metadata and an extensive set of ~40 million model-generated responses spanning a range of model capability levels. The graded difficulty design facilitates tailored training for models of varying sizes and capabilities, and also supports further applications such as downstream alignment methods including DPO\cite{rafailov2023direct}, GRPO\cite{shao2024deepseekmath}, and others. We release this dataset publicly to serve as a challenging benchmark and versatile training resource for the community.

\item Large-scale difficulty-aware reasoning training – We observe that enhancing the reasoning capabilities of base models, compared to traditional post-training methods, results in a training pattern shift, often requiring higher learning rates for effective training. By carefully selecting and scaling the difficulty of the training data, our reasoning-focused fine-tuning approach allows base models to significantly outperform existing distilled models on benchmarks like AIME2024, achieving near state-of-the-art performance.

\end{itemize}

We conclude by open-sourcing our entire methodology to facilitate further research. We have released our data to encourage transparency and rapid progress in building reasoning-strong LLMs from scratch. By sharing our ~40 million model-generated dataset and training recipe, we hope to enable the community to explore difficulty-aware training at scale and we hope this dataset empowers the research community to accelerate breakthroughs toward reasoning-strong LLMs.

\section{Data}
\label{section_data}


\subsection{Data Collection and Categories Definition}

To ensure comprehensiveness and diversity in our experimental data, we extensively collected datasets from multiple publicly available open-source corpora, clearly categorizing and defining each source. These datasets span tasks including mathematical reasoning, code generation, scientific reasoning, instruction-following, multi-turn conversations, and other/general reasoning.

The collected data were explicitly classified into the following categories:

\paragraph{Mathematical Reasoning}

Datasets in this category demand advanced numerical logic and reasoning capabilities from the model. We included datasets such as OpenR1-Math-220k\citep{openr1}, Big-Math-RL-Verified\citep{albalak2025bigmathlargescalehighqualitymath}, data\_ablation\_full59K\citep{muennighoff2025s1simpletesttimescaling}, NuminaMath\citep{numina_math_datasets}, MetaMathQA\citep{yu2023metamath}, 2023\_amc\_data\citep{aops2023amc8p10}, DeepMath-103K\citep{deepmath}, and AIME\_1983\_2024\citep{di_zhang_2025}\footnote{Note that we have excluded the problems from AIME2024.}.

\paragraph{Code Generation}

This category evaluates the model's ability in programming problem-solving and code generation tasks. Datasets selected include PRIME\citep{yuan2024implicitprm}, DeepCoder\citep{deepcoder2025}, KodCode\citep{xu2025kodcode}, liveincode\_generation\citep{jain2024livecodebench}, codeforces\_cots\citep{penedo2025codeforces}, verifiable\_coding\citep{openr1_verifiable_coding_2025}, opencoder\citep{Huang2024OpenCoderTO}, and OpenThoughts-114k-Code\_decontaminated\citep{openr1}, AceCode-87K\citep{AceCoder}.

\paragraph{Scientific Reasoning}

Scientific reasoning datasets primarily assess the model's performance in natural sciences and logical reasoning domains. We included datasets such as task\_mmmlu\citep{wang2022supernaturalinstructionsgeneralizationdeclarativeinstructions}, chemistryQA\citep{microsoft2021chemistryqa}, Llama-Nemotron-Post-Training-Dataset-v1\citep{nvidia2025llama3nemotron}, LOGIC-701\citep{hivaze_logic701_2023}, ncert\citep{NCERT_Physics_12th,NCERT_Physics_11th,NCERT_Chemistry_11th,NCERT_Chemistry_12th,NCERT_Biology_11th,NCERT_Biology_12th}, and logicLM\citep{longface2025logiclm}.

\paragraph{Instruction Following (IF)}

We selected four datasets closely related to instruction-following tasks, namely Llama-Nemotron-Post-Training-Dataset\citep{nvidia2025llama}, tulu-3-sft-mixture\citep{lambert2024tulu3}, if-eval-like, and AutoIF. Specifically, the if-eval-like dataset is aggregated from multiple data sources, and the AutoIF dataset was generated by leveraging the Qwen2.5-72B-Instruct\citep{qwen2, qwen2.5}. These datasets emphasize the model's ability to accurately comprehend and execute given instructions.

\paragraph{Multi-turn Conversations}

Multi-turn conversation datasets focus on maintaining context coherence and logical consistency across multiple interactions. Included datasets are InfinityInstruct\citep{InfinityInstruct2024, zhao2024iidoptimizinginstructionlearning, zhang2024inifinitymath}, OpenHermes-2.5\citep{OpenHermes_2_5}, tulu-3-sft-mixture\citep{lambert2024tulu3}, and ultra\_chat\citep{ding2023enhancing}.

\paragraph{Others/General Reasoning}

Datasets in this category cover a wide and diverse range of reasoning tasks, including open-ended queries, general knowledge reasoning, and everyday logical reasoning. The datasets selected here are evol\citep{wizardlm_evol_instruct_70k}, InfinityInstruct\citep{InfinityInstruct2024, zhao2024iidoptimizinginstructionlearning, zhang2024inifinitymath}, open\_orca\citep{OpenOrca}, tulu-3-sft-mixture\citep{lambert2024tulu3}, natural\_reasoning\citep{yuan2025naturalreasoningreasoningwild28m}, flan\citep{goodson2023huggyflan,longpre2023flan,wei2022finetuned,sanh2022multitask,wang2022supernaturalinstructions}, and OpenHermes-2.5\citep{OpenHermes_2_5}.

The chosen datasets and their defined categories comprehensively cover various task domains, enabling us to thoroughly assess the generalizability and effectiveness of our proposed data distillation method.

\subsection{Query Processing}

To ensure high quality and effectiveness of the data for subsequent model training, we applied meticulous preprocessing procedures to the collected raw queries. These procedures encompassed deduplication, filtering, and decontamination steps to mitigate redundancy and data contamination, thereby preventing negative impacts on the experimental outcomes.

\subsubsection{Deduplication and Filtering}
Initially, we conducted exact deduplication by removing queries with identical text. Furthermore, to enhance the data quality, stringent filtering criteria were employed as follows:

\begin{itemize}
    \item \textbf{Unicode Ratio Filtering}: Queries containing a high proportion of Unicode characters exceeding a predefined threshold were removed to avoid the introduction of meaningless or corrupted information.

    \item \textbf{Incomplete Query Filtering}: All queries that were empty or incomplete were removed to ensure query validity and completeness within the dataset.

    \item \textbf{Special Content Filtering}: Queries containing hyperlinks (URLs) or tabular content were excluded, as such content typically should not appear in training queries and could potentially cause hallucinations or misleading outputs from the model.
\end{itemize}

By implementing these precise deduplication and filtering strategies, we significantly improved the purity and quality of the dataset, laying a robust foundation for stable model training.

\subsubsection{Decontamination}

To prevent information leakage and semantic overlap between training data and the evaluation set, we performed rigorous decontamination, particularly focusing on the core evaluation dataset, AIME2024.

\begin{itemize}
    \item \textbf{Exact Matching Filtering}: We strictly filtered out queries that exactly matched or closely resembled any query in the AIME2024 evaluation set, ensuring the integrity and reliability of evaluation results.

    \item \textbf{Semantic Deduplication}: To further reduce implicit semantic overlap, we utilized semantic embedding techniques, specifically employing the bge-m3\cite{bge-m3} embedding model to calculate semantic similarity between training queries and the AIME2024 evaluation set. Queries with semantic similarity scores exceeding a threshold of 0.9 were identified as potentially contaminated and subsequently removed.
\end{itemize}

Through these rigorous exact matching and semantic deduplication processes, we substantially mitigated the risks of semantic conflict and information leakage, thereby ensuring that the evaluation results accurately reflect the model's generalization capabilities and true reasoning performance. Ultimately, we obtained approximately 3.34 million queries for further model training.

\subsection{Data Distilling}

To enhance data quality and accurately assess the difficulty of queries, we conducted data distillation using three models with progressively increasing capabilities: DeepSeek-R1-Distill-Qwen-1.5B, DeepSeek-R1-Distill-Qwen-7B, and DeepSeek-R1\citep{deepseekai2025deepseekr1incentivizingreasoningcapability}. Specifically, each of the approximately 4 million preprocessed queries was independently distilled four times by each model, generating corresponding reasoning processes and final answers. This resulted in approximately 40 million distilled responses in total. This approach allowed us to effectively capture variations among model outputs, facilitating subsequent computation of query difficulty.

It is important to note that although we completed distillation using all three models, due to constraints on time and computational resources, subsequent experimental analyses in this paper exclusively utilized the distilled responses from the DeepSeek-R1 model\footnote{Data from DeepSeek-R1-Distill-Qwen-1.5B, DeepSeek-R1-Distill-Qwen-7B, and DeepSeek-R1 have been publicly released. However, in this paper, only responses from the DeepSeek-R1 model were used for training and analysis.}.

\subsection{Ground Truth Verification}

To ensure the high quality of distilled data, we designed a rigorous and comprehensive ground truth verification process. Specifically, we adopted distinct verification methods tailored for different data categories, calculating verification scores (\textit{verify\_score}) accordingly, and subsequently derived the model's pass rate (\textit{pass\_rate}).

\subsubsection{Verification Score Calculation}

\paragraph{Mathematical Reasoning}
We employed a two-stage validation mechanism for mathematical data. Initially, results generated by models were assessed using Math-Verify\footnote{https://github.com/huggingface/Math-Verify}, yielding a binary outcome (correct as 1, incorrect as 0). If an initial result was identified as incorrect, a second validation was conducted using the Qwen2.5-7B-Instruct\citep{qwen2, qwen2.5}, again producing a binary outcome. The final \textit{verify\_score} was computed as:
\begin{equation}
\textit{verify\_score}_{\textit{math}} = \frac{\textit{Correct}_{\textit{Math-Verify}} + \textit{Correct}_{\textit{Second-Validation}}}{N_{\textit{total}}}
\end{equation}
where $N_{\textit{total}}$ is the total number of samples.

\paragraph{Code Generation}
For code generation data, the \textit{verify\_score} was calculated based on sandbox fusion tests using selected test cases. Specifically, the first 10 available open-source test cases were utilized per query, or all available test cases if fewer than 10 existed. Python codes were tested using standard input-output and assert-based formats, while C++ codes utilized only standard input-output tests. The \textit{verify\_score} was computed as follows:
\begin{equation}
\textit{verify\_score}_{\textit{code}} = \frac{\textit{Count}(\textit{test\_cases}_{\textit{passed}})}{\textit{Count}(\textit{test\_cases})}
\end{equation}

\paragraph{Scientific Reasoning}
Scientific reasoning data were validated using the Qwen2.5-7B-Instruct\citep{qwen2, qwen2.5}. The model assessed the similarity between generated results and ground truth, yielding decimal scores in the range [0, 5].

\paragraph{Instruction Following}
Verification for instruction-following (IF) data was performed using the ifeval validator. Due to limited constraint annotations in existing open-source datasets, we leveraged Qwen2.5-72B-Instruct to generate additional constraints based on queries. These constraints were formatted appropriately for ifeval validation, with the mean pass rate across all constraints used as the \textit{verify\_score}:
\begin{equation}
\textit{verify\_score}_{\textit{IF}} = \frac{\sum_{i=1}^{m} \textit{pass\_score\_constraint}_i}{m}, \quad \textit{pass\_score\_constraint}_i \in \{0,1\}
\end{equation}

Here, $m$ represents the total number of constraints per IF data point.

\paragraph{Multi-turn Conversations and Others}
For multi-turn conversations and general reasoning tasks, we employed Decision-Tree-Reward-Llama-3.1-8B\cite{rlhflow2025decisiontree} to evaluate three dimensions: coherence, correctness, and helpfulness, each scored in the range [0,4]. The composite reward score, serving as the \textit{verify\_score}, was computed as:
\begin{equation}
\textit{verify\_score}_{\textit{others}} = \frac{\textit{coherence} + \textit{correctness} + \textit{helpfulness}}{12}
\end{equation}

\subsubsection{Pass Rate Calculation}

Based on the calculated \textit{verify\_score}, we further determined each model's pass rate (\textit{pass\_rate}). We set strict thresholds for pass rate evaluation as follows:

\begin{itemize}
    \item Mathematical reasoning, code generation, and instruction-following data: \textit{verify\_score} > 0.99 indicates a pass.
    \item Scientific reasoning data: \textit{verify\_score} > 4.99 indicates a pass.
    \item Multi-turn conversations and other data: \textit{verify\_score} > 0.7 indicates a pass.
\end{itemize}

The pass rate per query was calculated as the average of binary pass indicators across four distillation attempts:
\begin{equation}
\textit{pass\_rate} = \frac{\sum_{i=1}^{n} \mathbf{1}(\textit{verify\_score}_i > \textit{threshold})}{n}, \quad n = 4
\end{equation}

\subsection{Quality Assurance}
In addition to the aforementioned data processing procedures, we introduced a series of supplementary filtering and quality assurance measures aimed at minimizing noise and anomalies in the dataset. These measures include:

\paragraph{Perplexity (ppl)-Based Filtering}
We utilized our previously trained 32B model\cite{zhao20251} to compute perplexity (ppl) scores for concatenated query and model-generated answers. Empirically, high perplexity scores indicate lower semantic coherence or logical consistency. Consequently, we set a threshold of 20, removing data instances with perplexity scores exceeding this value.

\paragraph{High-Frequency Ngram Filtering}
To prevent template-based or repetitive content, we applied Ngram frequency analysis across the entire dataset. Specifically, strings of 20 tokens that appeared more than 20 times were identified and removed as repetitive or redundant content. This step significantly enhanced the diversity and originality of the distilled data.

\paragraph{Additional Logical and Structural Checks}
Beyond statistical filtering, we conducted structural and logical validations to ensure data completeness and consistency, including:
\begin{itemize}
    \item \textbf{Even Turn Verification}: For multi-turn conversation datasets, we required an even number of dialogue turns to maintain contextual integrity and coherence.
    \item \textbf{Think Content Verification}: All distilled data instances were required to explicitly contain intermediate reasoning steps (think content). Instances lacking such content were considered invalid and excluded.
    \item \textbf{Answer Content Verification}: All data instances needed to contain a clear and complete final answer (answer content). Instances without explicit answers were considered incomplete and removed.
\end{itemize}

For details on data processing procedures and additional information, please refer to Appendix \ref{data_analysis}.

\section{Methodology}
\label{section_methodology}


In this section, we aim to identify and scale the most valuable training samples to progressively enhance the model's learning capability. We first introduce the definition of the Coefficient of Variation (CV) and then elaborate on our data selection procedure. Additionally, we conduct preliminary experiments using a two-stage annealing training strategy to further improve model performance.

\subsection{Coefficient of Variation Definition}
\label{section_data_selection}

Specifically, for each query, we first generated $n$ independent responses using DeepSeek-R1 and computed a \textit{verify\_score} for each response (See Section \ref{section_data}), measuring its correctness or quality. However, not all generated data contribute equally to model training. Queries with extremely high average \textit{verify\_score} indicate that these questions are too simple for the model, while queries with very low average scores might indicate that these problems are too difficult or inadequately learned by the model. Therefore, queries with extreme average scores should be excluded, allowing us to focus on queries with higher learning potential.

Moreover, the variability of the \textit{verify\_score} across multiple responses to the same query is crucial. High variability indicates instability in the model's performance, highlighting significant room for improvement. Thus, we employ the \textit{Coefficient\ of\ Variation (CV)} as the key indicator to quantify this variability.

The \textit{Coefficient of Variation} is defined as follows:
\begin{equation}
\textit{CV} = \frac{\sqrt{\frac{1}{n}\sum_{i=1}^{n}(\textit{verify\_score}_i - \mu)^2}}{\mu} = \frac{\sigma}{\mu}, \quad \textit{where} \ \mu = \frac{1}{n}\sum_{i=1}^{n}\textit{verify\_score}_i
\end{equation}

The $CV$ is a normalized dimensionless metric, objectively reflecting the variability of data.

For example, consider two queries with the following \textit{verify\_score} values:

\begin{align}
&Q_A: [0.5, 0.5, 0.5, 0.5, 0.5], \mu = 0.5, \sigma = 0, CV = 0 \notag \\
&Q_B: [0.9, 0.1, 0.7, 0.3, 0.5], \mu = 0.5, \sigma \approx 0.32, CV \approx 0.63 \notag
\end{align}

Although both queries have the same mean score, $Q_B$ exhibits significantly higher variability (higher $CV$), indicating that the model's performance on $Q_B$ is less stable, hence $Q_B$ has greater learning value than $Q_A$.

Consider another example with two queries:

\begin{align}
&Q_C: [0.6, 0.8, 0.7, 0.7, 0.7], \mu = 0.7, \sigma \approx 0.07, CV \approx 0.10 \notag \\
&Q_D: [0.3, 0.5, 0.4, 0.4, 0.4], \mu = 0.4, \sigma \approx 0.07, CV \approx 0.18 \notag
\end{align}

Despite having equal standard deviations (similar variability), $Q_D$ has a lower mean, suggesting that the model's performance on $Q_D$ is inferior, making $Q_D$ more valuable for learning.

By employing the $CV$-based selection strategy, we effectively identify the most beneficial data for model improvement, thus significantly enhancing the quality of the training dataset and the model's generalization capabilities.

\subsection{Data Selection}

\paragraph{Large Scale Reasoning Data Selection (Stage I)}
\label{section_stage_1}

We filter data based on the \textit{verify\_score} (defined in Section~\ref{section_data}) and the \textit{Coefficient of Variation (CV)} (defined in Section~\ref{section_data_selection}) for each query, using multiple responses generated by the model. This filtering aims to ensure the effectiveness and quality of the training dataset.

\begin{algorithm}[H]
\caption{Data Filtering Procedure}
\label{alg:stage1_filter}
\begin{algorithmic}[1]
\Require A set of $n$ responses per query with corresponding \textit{verify\_score}
\Require Category-specific \textit{verify\_score threshold} and fixed \textit{CV threshold} ($CV_t$)
\Ensure Filtered high-quality response set

\For{each query $q$ in dataset}
    \State Compute the maximum \textit{verify\_score} $s_{\max}$
    \If{$s_{\max} < \textit{verify\_score threshold for } q$}
        \State Discard query $q$
        \State \textbf{continue}
    \EndIf
    \State Compute \textit{Coefficient of Variation} (CV) for $q$
    \If{CV $>$ $CV_t$}
        \Comment{\textbf{Challenging query}}
        \State Retain only responses with \textit{verify\_score} $>$ threshold
    \Else
        \Comment{\textbf{Easy query}}
        \If{Category of $q$ is \textit{other} or \textit{multiturn}}
            \State Keep with probability 0.5
        \Else
            \State Discard query $q$
        \EndIf
    \EndIf
\EndFor
\State \Return Filtered dataset
\end{algorithmic}
\end{algorithm}

Specifically, we define two key selection criteria: \textit{maximum verify\_score threshold} and \textit{CV threshold}. For all data categories, we apply strict \textit{maximum verify\_score threshold} to guarantee response correctness. For categories with standardized answers, such as \textit{math}, \textit{code}, and \textit{instruction follow}, we set the \textit{maximum verify\_score threshold} to 0.99, and 4.99 for \textit{science}. For more complex or multi-turn conversational data (\textit{other} or \textit{multiturn}), the threshold is relaxed to 0.7 to accommodate the inherent difficulty and variability of such queries.

\begin{figure}[ht]
    \centering
    \includegraphics[width=0.95\linewidth]{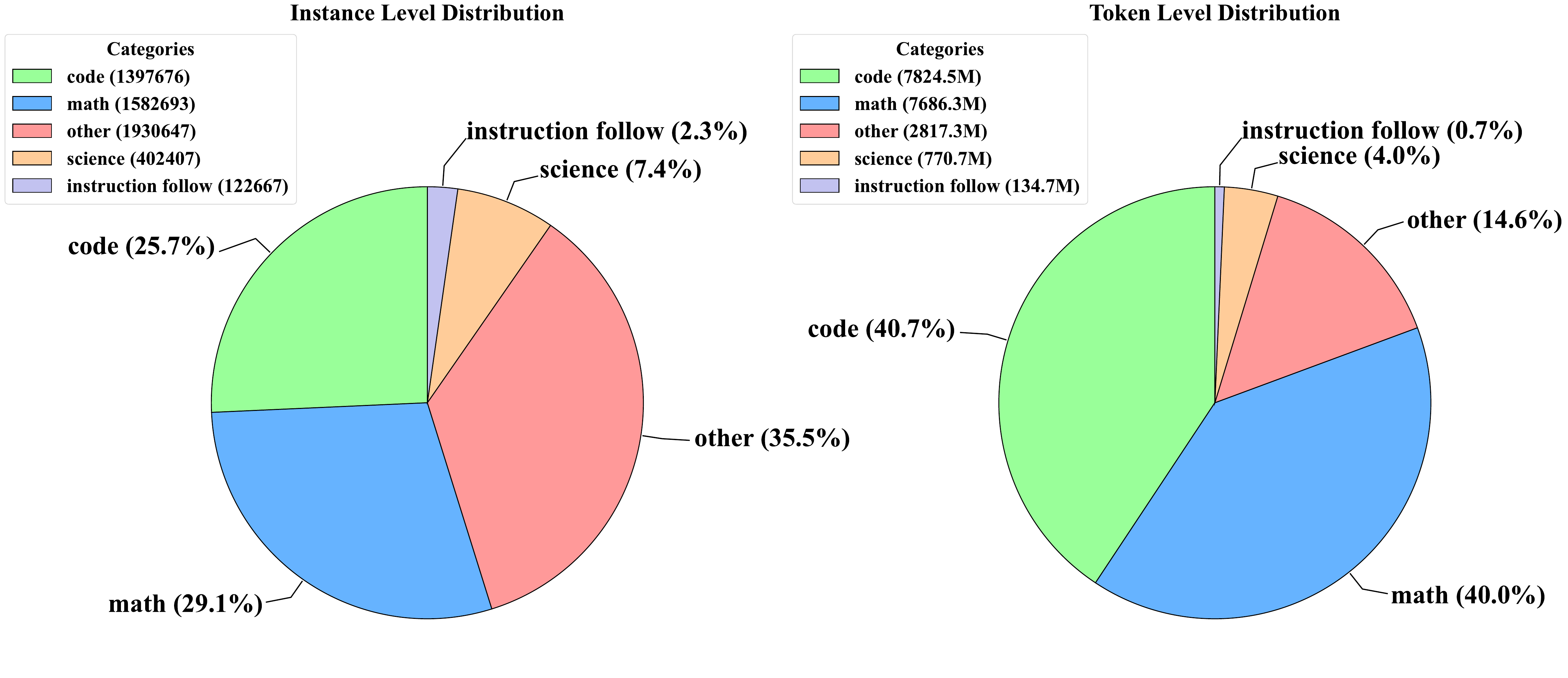}
    \caption{Distribution of training data types in Supervised Fine-Tuning (SFT) Stage I. The left pie chart illustrates the proportion at the instance-level, while the right pie chart shows the distribution at the answer token-level.}
    \label{fig:stage_1_combined_category_distribution}
\end{figure}

The filtering process is as follows:

\begin{enumerate}[label=\arabic*)]
    \item First, we evaluate whether the highest \textit{verify\_score} among the $n$ responses for each query exceeds the corresponding category-specific threshold. If not, the query is discarded, as it indicates limited learning potential due to poor model performance.
    \item If the highest \textit{verify\_score} passes the threshold, we further assess the variability using the \textit{CV}:
    \begin{enumerate}[label=\alph*)]
        \item If the \textit{CV} exceeds the threshold of $CV_t$\footnote{Note that different category has different \textit{CV threshold}.}, the model's performance on this query is considered unstable, suggesting that the query is challenging and holds high learning potential. In this case, only the responses that exceed the \textit{verify\_score} threshold are retained to ensure data quality.
        \item If the \textit{CV} is less than or equal to $CV_t$, the query is deemed easy, and the model's performance is stable. To maintain diversity and learnability within the dataset, we retain these queries based on their category. For \textit{other} and \textit{multiturn} data, we retain 50\% of the samples at random to avoid excessive simplification of the dataset; for all other categories, such queries are discarded by default.
    \end{enumerate}
\end{enumerate}

Through this carefully designed filtering strategy, we successfully curated a high-quality dataset consisting of \textbf{5 million} reasoning-intensive samples with strong learning value.

\paragraph{Exploration of Annealed Data Selection (Stage II)}
\label{sec_stage_2}

Previously, we initially filtered high-quality and moderately challenging data by jointly considering the \textit{verify\_score} and the \textit{Coefficient of Variation (CV)}. However, as the model significantly improves its reasoning capabilities after large scale reasoning data training, the most beneficial data for the model in annealing stage consists of queries that are inherently more complex, uncertain, and challenging. Therefore, in the annealing stage, we specifically focus on further increasing the difficulty of the training data by exclusively selecting queries with higher variability, aiming to maximize the model's learning effectiveness.

\begin{figure}[ht]
    \centering
    \includegraphics[width=0.95\linewidth]{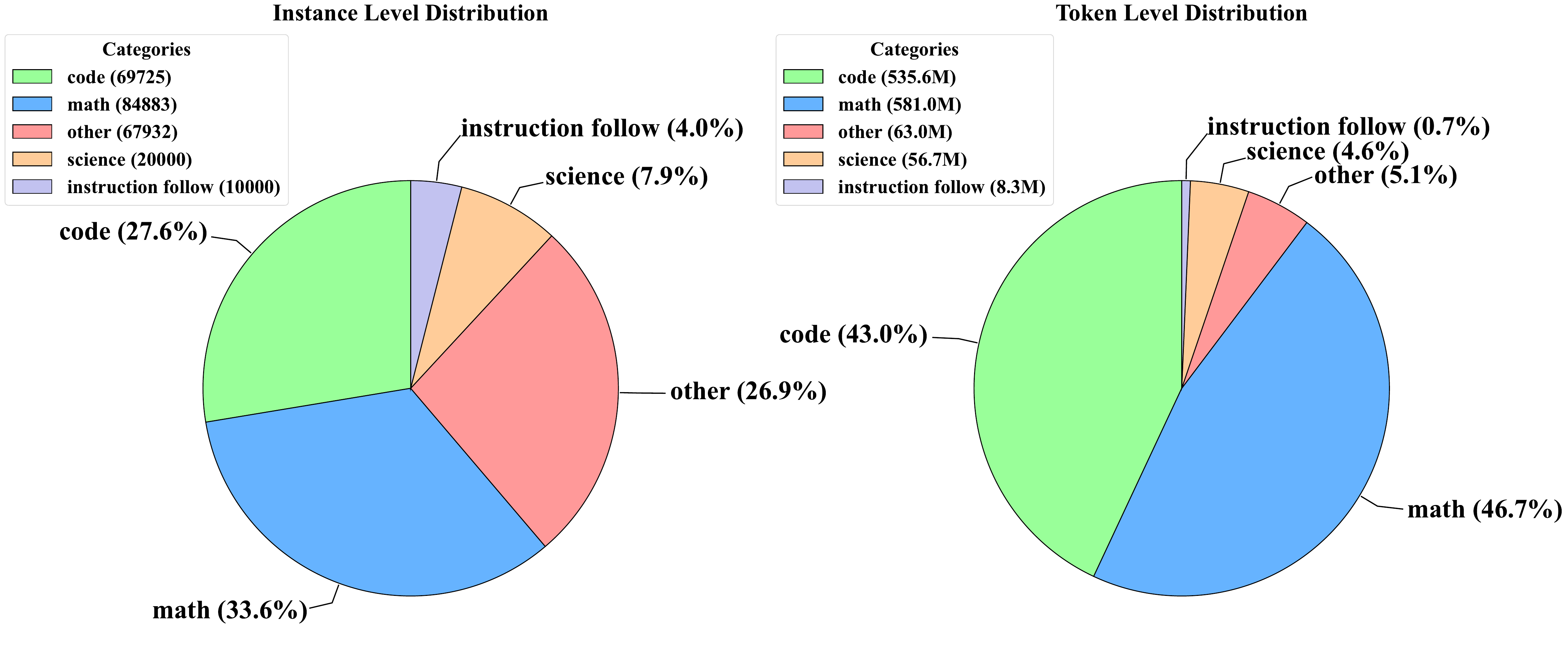}
    
    \caption{Distribution of training data types in Supervised Fine-Tuning (SFT) Stage II. The left pie chart illustrates the proportion at the instance-level, the right pie chart shows the distribution at the answer token-level.}
    \label{fig:stage_2_combined_category_distribution}
\end{figure}

Specifically, we maintain the same categories as defined in Stage I (\textit{math}, \textit{code}, \textit{science}, \textit{instruction follow}, and \textit{other}), set a stringent \textit{verify\_score threshold} of 0.99, retaining only those queries exceeding $CV_t$. Moreover, we exclusively select queries exhibiting the highest CV values, and for multiple model responses to a single query, we randomly sample only one response for training.

Through this rigorous data selection strategy, we extract nearly 200k challenging and highly reliable training data, enabling the model to improve further in more difficult reasoning scenarios.

\section{Experiments}

\subsection{Large Scale Reasoning Training}

In the large scale reasoning training stage, we trained the model using the high-quality dataset obtained through the previously described data selection strategy (see Section~\ref{section_stage_1}). Specifically, we employed a learning rate of $8\times10^{-5}$ and trained the model for one epoch. To optimize computational resources and improve training efficiency, we utilized a packing strategy, grouping sequences up to a maximum length of 32k tokens, with a global batch size of 64. Additionally, we applied a cosine learning rate scheduler with a linear warmup phase for the initial 5\% of the training steps, subsequently decaying the learning rate to zero.


Compared to traditional post-training methods, we observed a training pattern shift when enhancing reasoning abilities based on base models. Specifically, a relatively high learning rate (e.g., $8\times10^{-5}$ as adopted in our experiments) becomes crucial to effectively capture complex data features. Lower learning rates often led to underfitting, limiting the model's ability to sufficiently grasp long-sequence reasoning tasks. Therefore, we employed a higher learning rate during our data selection process to facilitate the model's learning in complex reasoning scenarios. We conducted experiments using Qwen-2.5-32B and Qwen-2.5-72B models\cite{qwen2, qwen2.5}.

\subsection{Annealing}
We performed an annealing training experiment starting from the optimal 32B model obtained previously. Unlike Stage I, we adopted a more stringent data selection strategy, retaining only the highly challenging data characterized by a coefficient of variation (\textit{CV}) greater than $CV_t$ (see Section \ref{sec_stage_2}). In this stage, we discontinued the use of the packing strategy to allow the model to focus exclusively on handling complex long-reasoning tasks for individual query-response pairs, while still maintaining a maximum sequence length of 32k tokens. We set a lower learning rate of $8\times10^{-6}$ and trained the model for two epochs, further exploring the model's potential on challenging long-reasoning tasks. Similarly, we employed a cosine learning rate scheduler with a linear warmup phase for the first 5\% of the training steps, gradually decaying the learning rate to zero.

\subsection{Benchmarks}



To rigorously evaluate the effectiveness of our approach, we conducted experiments on several representative benchmarks, including AIME2024\citep{maa_aime_2024}, LiveCodeBench\citep{jain2024livecodebench}, and GPQA-Diamond\citep{rein2023gpqagraduatelevelgoogleproofqa}. Specifically, we utilized pass@1 for AIME2024, LiveCodeBench, and GPQA-Diamond, consistent with previous studies.

Among these benchmarks, AIME2024 is a challenging mathematical reasoning competition dataset comprising 30 integer-answer questions designed to assess precise mathematical reasoning; LiveCodeBench is a comprehensive, contamination-free coding benchmark, continuously aggregating new programming challenges from platforms such as LeetCode, AtCoder, and Codeforces; GPQA-Diamond is a subset of the GPQA dataset containing 198 high-difficulty graduate-level multiple-choice questions developed by experts in biology, physics, and chemistry to evaluate advanced scientific reasoning.

\subsection{Results and Analysis}

\subsubsection{Overall Results}
\begin{table}[htbp]
    \caption{Performance comparison of various models.}
    \label{tab:model_performance}
    \centering
    \renewcommand{\arraystretch}{1.3} 
    \begin{tabular}{l c c c}
        \hline
        Model & AIME2024 (\%) & GPQA-Diamond (\%) & LiveCodeBench (\%) \\
        \hline
        DS-Distill-32B {\scriptsize (SFT)} & 72.6 & 62.1 & 57.2 \\
        QwQ-32B {\scriptsize (RL)} & 79.5 & 65.9 & 63.4 \\
        DS-Distill-70B {\scriptsize (SFT)} & 70.0 & 65.2 & 57.5 \\
        DeepSeek-R1 {\scriptsize (RL)} & 79.8 & 71.5 & 65.9 \\
        \hline
        Ours-Distill-32B {\scriptsize (SFT)} & 75.8 & 66.3 & 64.2 \\
        Ours-Distill-72B {\scriptsize (SFT)} & 79.2 & 65.7 & 63.8 \\
        \hline
    \end{tabular}
\end{table}

Table \ref{tab:model_performance} presents our experimental results. On AIME2024, our 72B model achieved a pass rate of 79.2\%, and our 32B model achieved 75.8\%, both significantly outperforming DeepSeek-Distill-32B (72.6\%) and DeepSeek-Distill-70B (70.0\%). Notably, despite relying solely on supervised fine-tuning (SFT), our 72B model closely approaches reinforcement learning-based models such as QwQ-32B (79.5\%) and DeepSeek-R1 (79.8\%). Additionally, our 32B model demonstrates superior performance over DeepSeek-Distill models on other evaluation benchmarks, achieving 66.3\% on GPQA-Diamond and 64.2\% on LiveCodeBench, indicating robust generalization and enhanced reasoning capabilities.

\subsubsection{Analysis}

\begin{table}[htbp]
    \caption{Model performance of two stage SFT models.}
    \label{tab:model_performance_2_stage}
    \centering
    \renewcommand{\arraystretch}{1.3} 
    \begin{tabular}{l c c c}
        \hline
        Model & AIME2024 (\%) & GPQA-Diamond (\%) & LiveCodeBench (\%) \\
        \hline
        Ours-32B {\scriptsize (Stage I)} & 75.8 & 66.3 & 64.2 \\
        Ours-32B {\scriptsize (Stage II)} & 77.9 & 66.9 & 61.7 \\
        \hline
    \end{tabular}
\end{table}

Comparisons between large scale reasoning training (32B Training Stage I) and annealing (32B Training Stage I) (Table \ref{tab:model_performance_2_stage}) indicate that our two-stage fine-tuning strategy improves performance on several key benchmarks. Specifically, on AIME2024, our 32B model improved from 75.8 (Stage I) to 77.9 (Stage II), an increase of 2.1 percentage points, and GPQA-Diamond improved from 66.3 to 66.9. However, performance on LiveCodeBench slightly decreased from 64.2 (Stage I) to 61.7 (Stage II), indicating that further hyperparameter tuning and exploration of the optimal mixing ratio for code-related data are necessary to consistently benefit from the two-stage training across all tasks.

\begin{figure}[ht]
    \centering
    \includegraphics[width=0.9\linewidth]{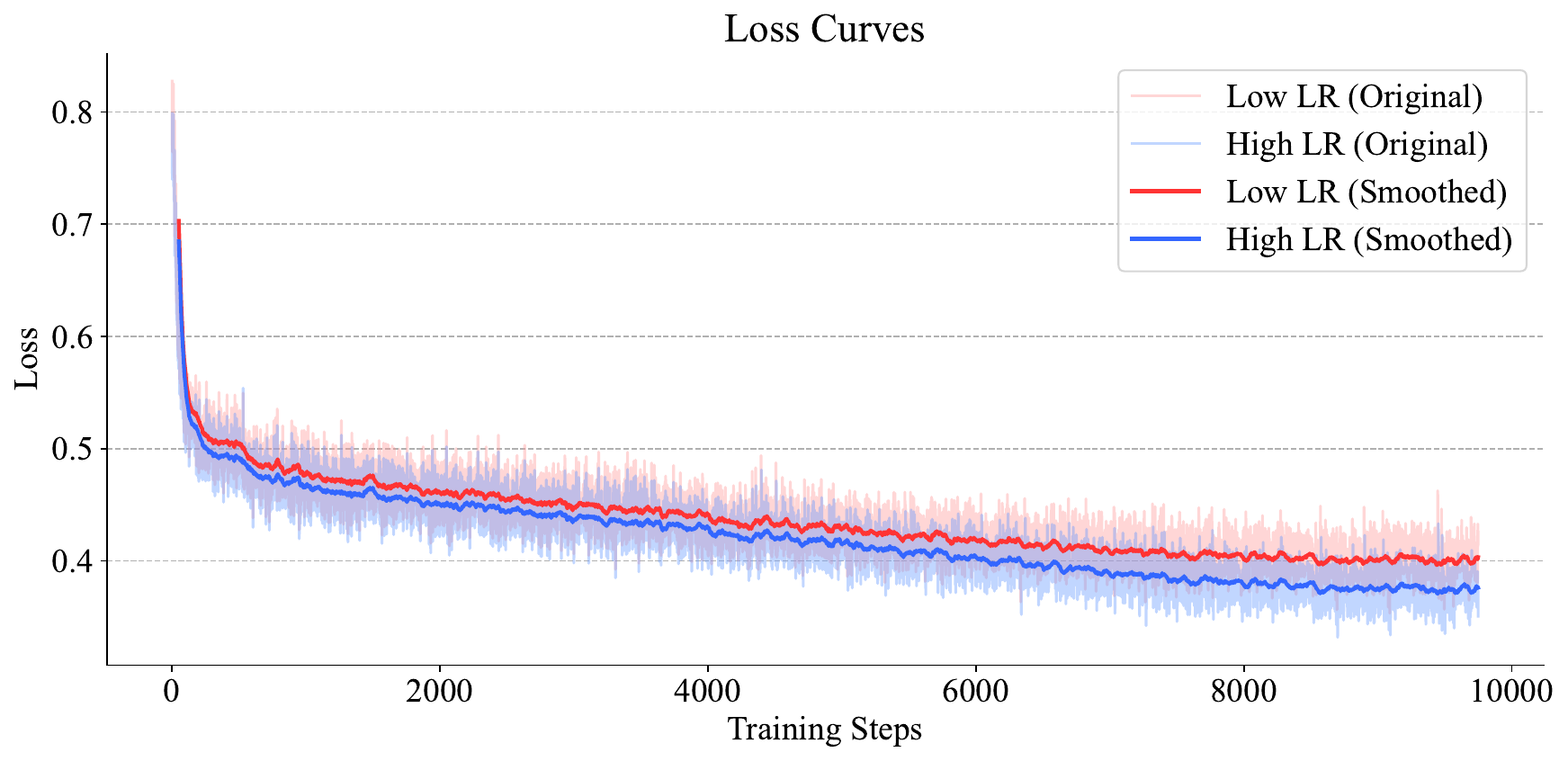}
    \caption{Loss curves of 72B model training.}
    \label{fig:loss_curve}
\end{figure}

We observed a training pattern shift when conducting reasoning-enhancement training based on the base model, compared to traditional post-training approaches. Specifically, reasoning-focused fine-tuning required a higher learning rate to adequately capture complex reasoning features present in the training data. To quantitatively illustrate this phenomenon, we conducted an ablation study using our 72B model. When reducing the learning rate from $8\times10^{-5}$ to $8\times10^{-6}$, we observed significant performance degradation across evaluation benchmarks. On the AIME2024 dataset, performance dropped by 6.7 percentage points (from 79.2\% to 72.5\%), while on LiveCodeBench, it declined by 3.6 percentage points (from 63.8\% to 60.2\%). As depicted in Figure \ref{fig:loss_curve}, the training loss curves clearly demonstrate that a higher learning rate allows the model to better fit complex reasoning data, highlighting the necessity of adjusting training strategies for enhanced reasoning capabilities.

\begin{figure}[ht]
    \centering
    \includegraphics[width=1\linewidth]{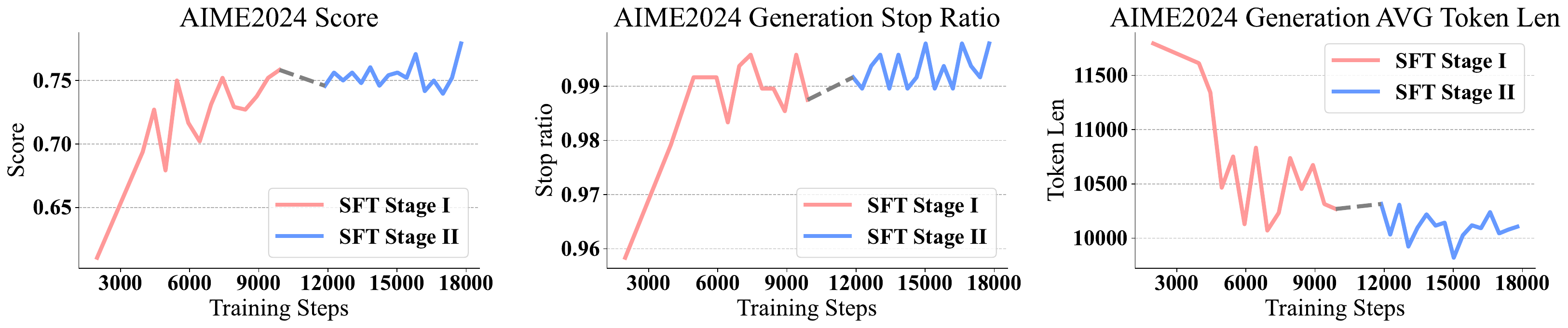}
    \caption{The variations of 32B AIME2024 Score, Generation Stop Ratio, and Average Generated Token Lengths with training steps}
    \label{fig:aime2024_metrics}
\end{figure}

Additionally, we analyzed the 32B model's performance metrics on AIME2024 as training steps increased (Figure \ref{fig:aime2024_metrics}). In Stage I, due to the larger dataset and higher learning rate, the model quickly learned and rapidly improved in performance. In Stage II, we employed a lower learning rate and omitted the packing strategy, enabling more refined learning and leading to further improvements in the final performance. Moreover, since foundational models are typically trained on vast amounts of text data rather than QA-specific tasks, initially the generation stop ratio was relatively low, indicating frequent failures in stopping generated outputs. However, with continuous training, the stop ratio significantly improved, and the average generated token length stabilized, demonstrating effective adaptation to the QA format.

Our difficulty-aware training approach substantially enhanced foundational model performance on complex reasoning tasks, closely approaching state-of-the-art proprietary model levels. This highlights the significant potential of using Supervised Fine-Tuning (SFT) alone on base models, demonstrating that applying suitable data and optimized training strategies can yield highly competitive results.

\section{Conclusion and Limitations}

In this paper, we introduce a large-scale, difficulty-graded reasoning dataset containing approximately 3.34 million unique queries with 40 million model-generated responses spanning various capability gradients. Leveraging this dataset, we conduct large-scale reasoning-focused training based on pass rates and the Coefficient of Variation (CV) to identify and utilize the most valuable training data. Additionally, we observed a training pattern shift, indicating that higher learning rates are necessary for effectively fitting complex reasoning tasks.

Moreover, preliminary experiments revealed that further increasing data complexity through an annealed second-stage training could enhance model performance. Starting from a base model, our reasoning-focused fine-tuning method surpasses most open-source distilled models on several complex reasoning benchmarks, such as AIME2024, achieving performance close to or at state-of-the-art levels. These results validate the effectiveness and generalization capability of both our dataset and training strategies.


In future work, we aim to develop more refined methods for evaluating data quality to better identify data most beneficial for model training. Additionally, we will investigate how models with varying initial capabilities influence subsequent reinforcement learning (RL) training outcomes, thus illuminating the interplay between foundational model capability and data quality. We hope that our open-sourced data and methodologies will facilitate further advancements in constructing open-source large language models with exceptional long-form reasoning capabilities, contributing significantly to the broader research community.


\bibliographystyle{plainnat}
\bibliography{reference}

\clearpage
\appendix

\section{Data Analysis}
\label{data_analysis}

\subsection{Data Processing Pipeline}

\begin{figure}[ht]
    \centering
    \includegraphics[width=0.975\linewidth]{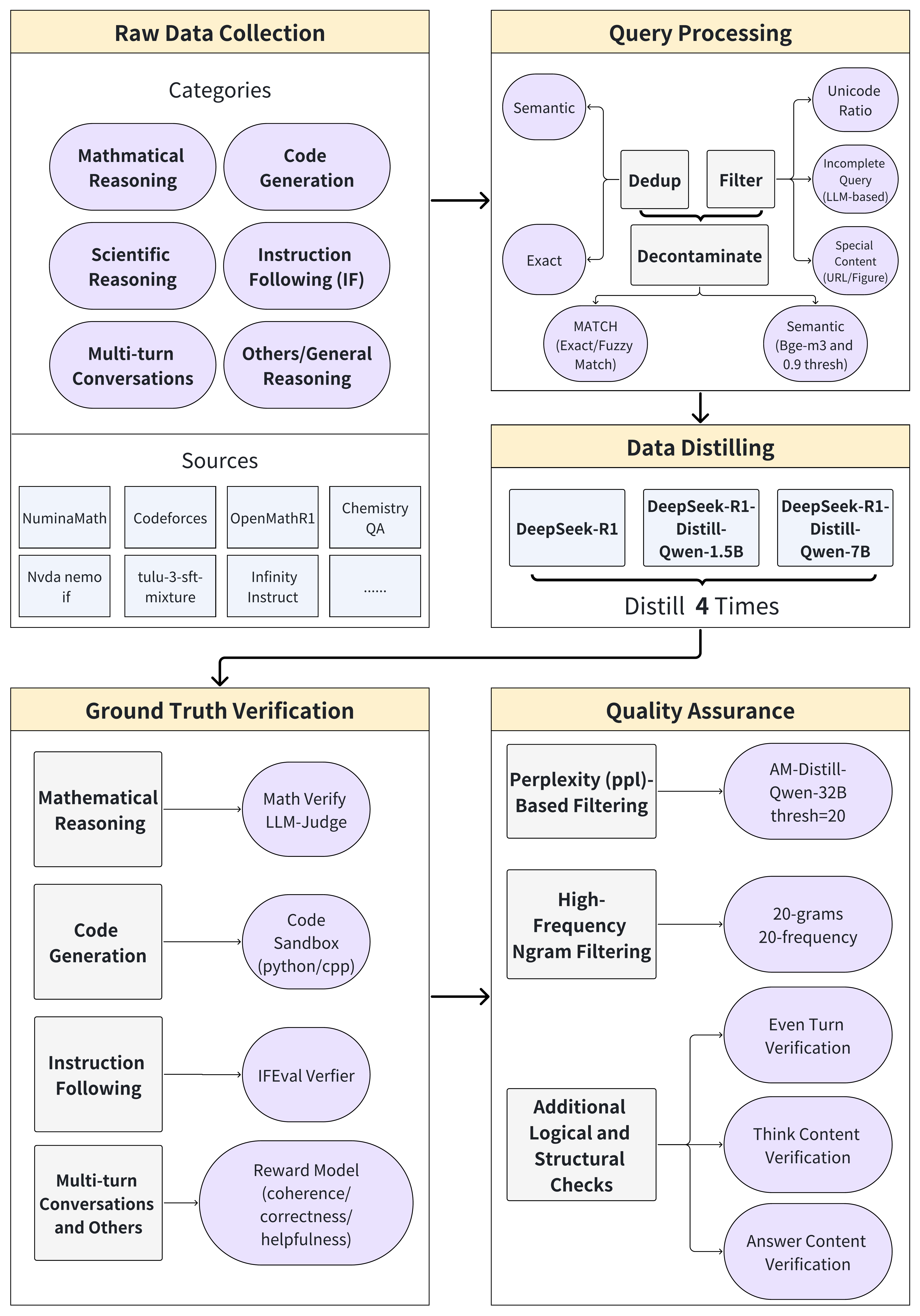}
    \caption{Construction process of data pipeline.}
    \label{fig:data_processing_pipeline}
\end{figure}

\subsection{Category Distribution}

\begin{figure}[ht]
    \centering
    \includegraphics[width=0.72\linewidth]{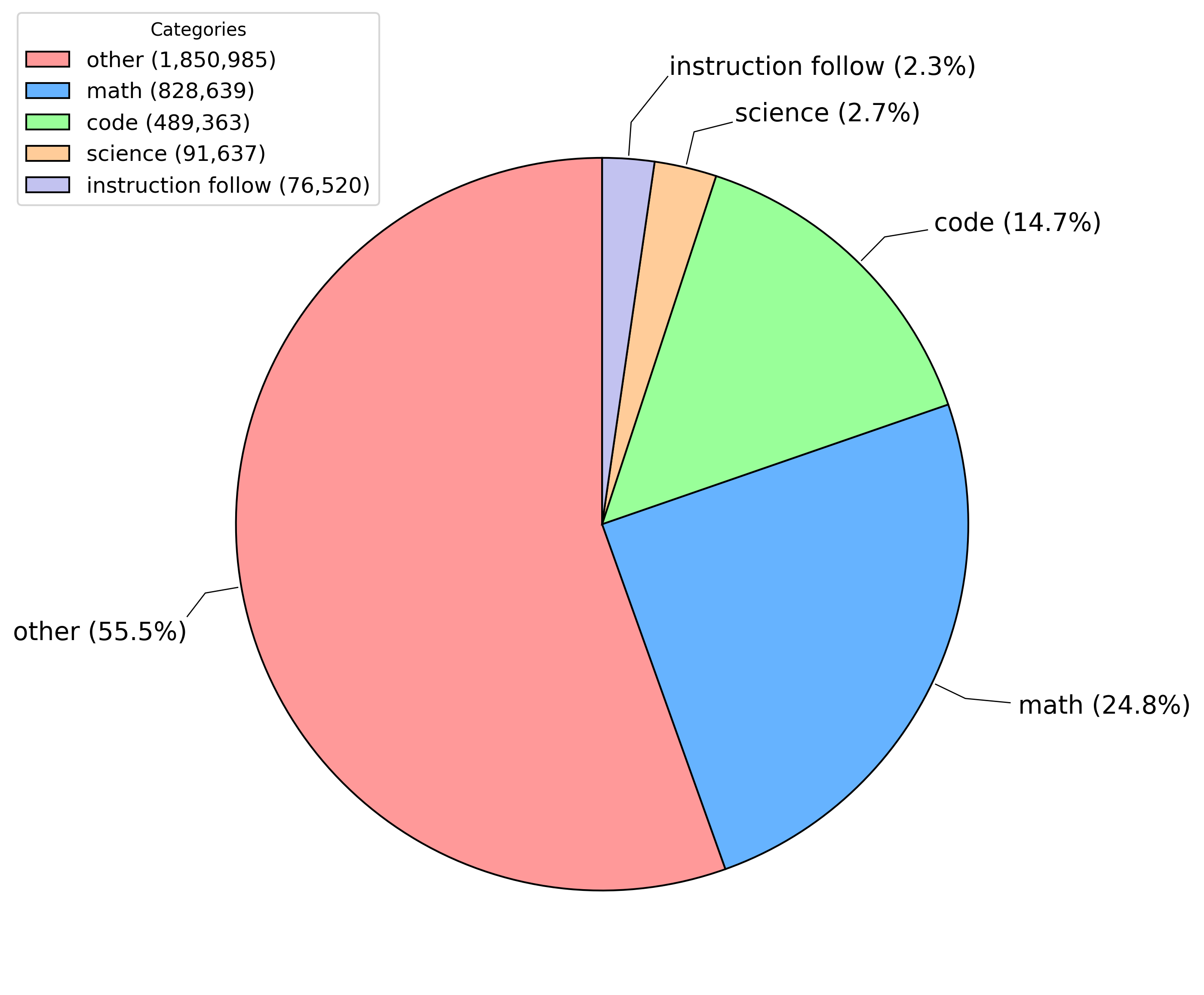}
    \caption{Distribution of query types in the dataset. Math queries constitute 24.8\%, code 14.7\%, science 2.7\%, instruction following 2.3\%, and other 55.5\%.}
    \label{fig:category_distribution}
\end{figure}

\subsection{Pass Rate Distribution}

\begin{figure}[ht]
    \centering
    \includegraphics[width=0.72\linewidth]{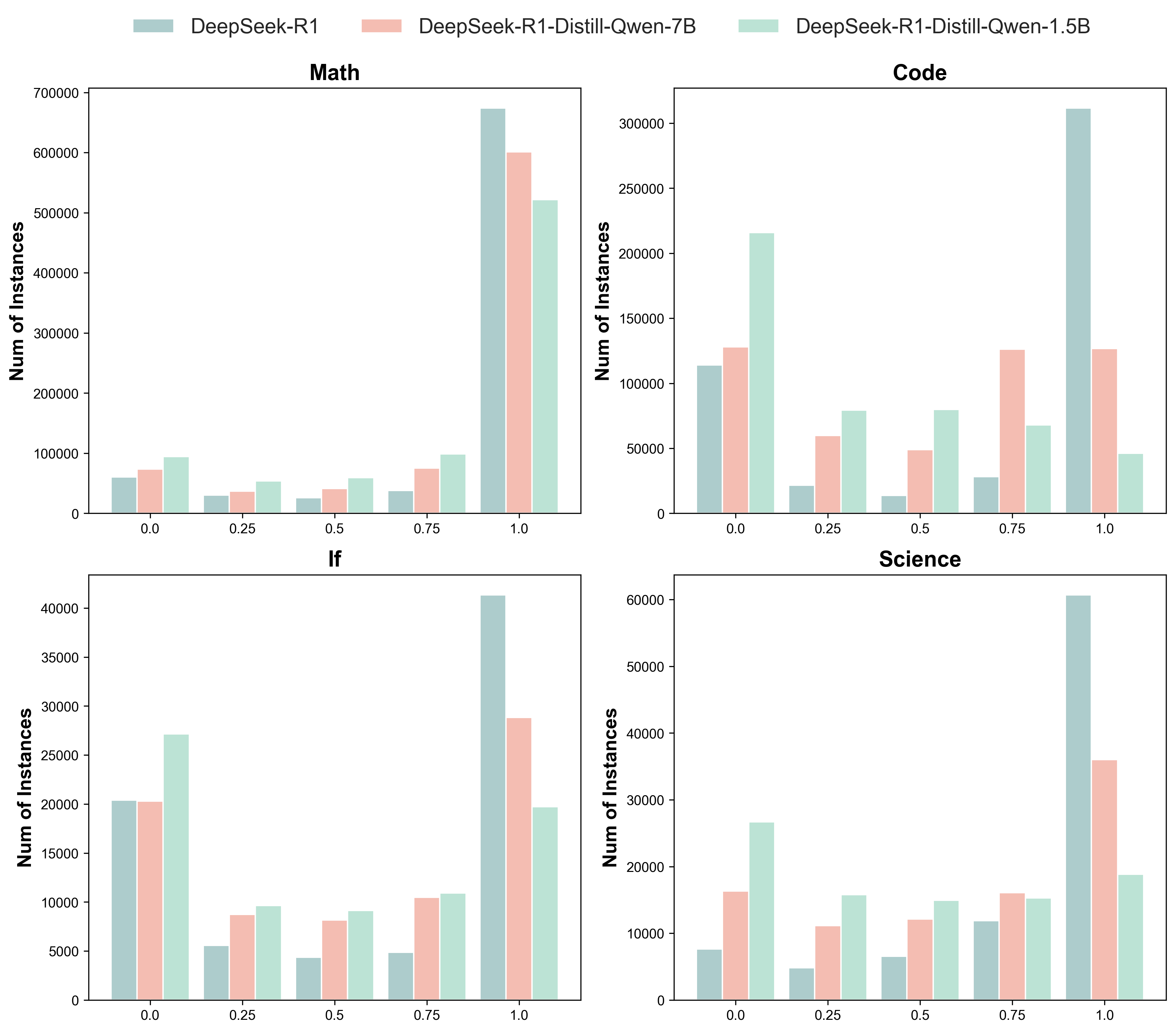}
    \caption{Pass rate distributions of the three models across four categories of data: math, code, science, and instruction following.}
    \label{fig:pass_rate_distribution}
\end{figure}


\end{document}